\documentclass[letterpaper, 10 pt, conference]{ieeeconf}  % Comment this line out if you need a4paper

\IEEEoverridecommandlockouts                              % This command is only needed if 
\usepackage{amsmath} % assumes amsmath package installed
\usepackage{amssymb}  % assumes amsmath package installed
\usepackage[strings]{underscore}
\usepackage{eqexpl}
\eqexplSetDelim{:}
\usepackage{multirow}
\usepackage[per-mode = symbol]{siunitx}   % physical quantities and SI units
\usepackage{booktabs}
\usepackage{graphicx}
\usepackage{cleveref}
\usepackage{algorithm}
\usepackage{algpseudocode}

\usepackage{float}
\usepackage{caption}
\usepackage{multicol}
\usepackage{makecell}
\DeclareSIUnit{\mph}{mph}
\usepackage{cite}
\newcolumntype{P}[1]{>{\centering\arraybackslash}m{#1}}
\newcolumntype{L}[1]{>{\leftjustified\arraybackslash}m{#1}}
\title{\LARGE \bf
Integrity Monitoring of 3D Object Detection in Automated Driving Systems using Raw Activation Patterns and Spatial Filtering}

\author{Hakan Yekta Yatbaz$^{1}$, Mehrdad Dianati$^{1,2}$, Konstantinos Koufos$^{1}$ and  Roger Woodman$^{1}$% <-this % stops a space
\thanks{*This research has been conducted as part of the EVENTS project, which is funded by the European Union, under grant agreement No 101069614. Views and opinions expressed are however those
of the author(s) only and do not necessarily reflect those of the European Union or European Commission. Neither the European Union nor the granting authority can be held responsible for them. This research has been also supported by the Centre for Doctoral Training (CDT) to Advance the Deployment of Future Mobility Technologies at the University of Warwick.}% <-this % stops a space
\thanks{$^{1}$Connected and Cooperative Autonomous Systems, WMG, University of Warwick, Coventry, United Kingdom
        {\tt\small \{hakan.yatbaz, m.dianati, k.koufos, r.woodman\}@warwick.ac.uk}}%
\thanks{$^{2}$School of Electronics, Electrical Engineering and Computer Science (EEECS), Queen’s University of Belfast, Belfast, United Kingdom
{\tt\small m.dianati@qub.ac.uk}}%
}

\begin{document}

\maketitle
\thispagestyle{empty}
\pagestyle{empty}

%%%%%%%%%%%%%%%%%%%%%%%%%%%%%%%%%%%%%%%%%%%%%%%%%%%%%%%%%%%%%%%%%%%%%%%%%%%%%%%%
\begin{abstract}
The deep neural network (DNN) models are widely used for object detection in automated driving systems (ADS). Yet, such models are prone to errors  which can have serious safety implications. Introspection and self-assessment models that aim to detect such errors are therefore of paramount importance for the safe deployment of ADS. Current research on this topic has focused on techniques to monitor the integrity of the perception mechanism in ADS. Existing introspection models in the literature, however, largely concentrate on detecting perception errors by assigning equal importance to all parts of the input data frame to the perception module. This generic approach overlooks the varying safety significance of different objects within a scene, which obscures the recognition of safety-critical errors, posing challenges in assessing the reliability of perception in specific, crucial instances. Motivated by this shortcoming of state of the art, this paper proposes a novel method integrating raw activation patterns of the underlying DNNs, employed by the perception module, analysis with spatial filtering techniques. This novel approach enhances the accuracy of runtime introspection of the DNN-based 3D object detections by selectively focusing on an area of interest in the data, thereby contributing to  the safety and efficacy of ADS perception self-assessment processes. 
\end{abstract}

%%%%%%%%%%%%%%%%%%%%%%%%%%%%%%%%%%%%%%%%%%%%%%%%%%%%%%%%%%%%%%%%%%%%%%%%%%%%%%%%
\section{Introduction}\label{sec:intro}
Perception is essential for automated driving systems (ADS) to understand their surroundings and safely navigate across various driving scenarios. A failure in perception may not only cause discomfort to the passengers but can also result in serious accidents. For example, the reports available in~\cite{eduardo1} and~\cite{eduardo2} discuss cases where the perception system of an automated vehicle failed to identify objects in its vicinity, causing fatal injuries to the passengers. Hence, it is crucial that ADS incorporate continuous monitoring mechanisms of their perception systems~\cite{hakansurvey, rahmansurvey} to robustly identify and handle runtime errors. In such cases, an alert must be issued, which can be used to hand over the control of the vehicle to the driver or trigger a minimum risk manoeuvre, depending on the automation level defined by SAE \cite{sae}.

%must trigger an alert. Depending on the level of autonomy, an alert can hand over the control back to the human driver in SAE level 3, or initiate a minimum risk manoeuvre in SAE level 4. % see Fig.~\ref{fig:overview}. 
\begin{figure}[t]
    \centering
    \includegraphics[width =\linewidth]{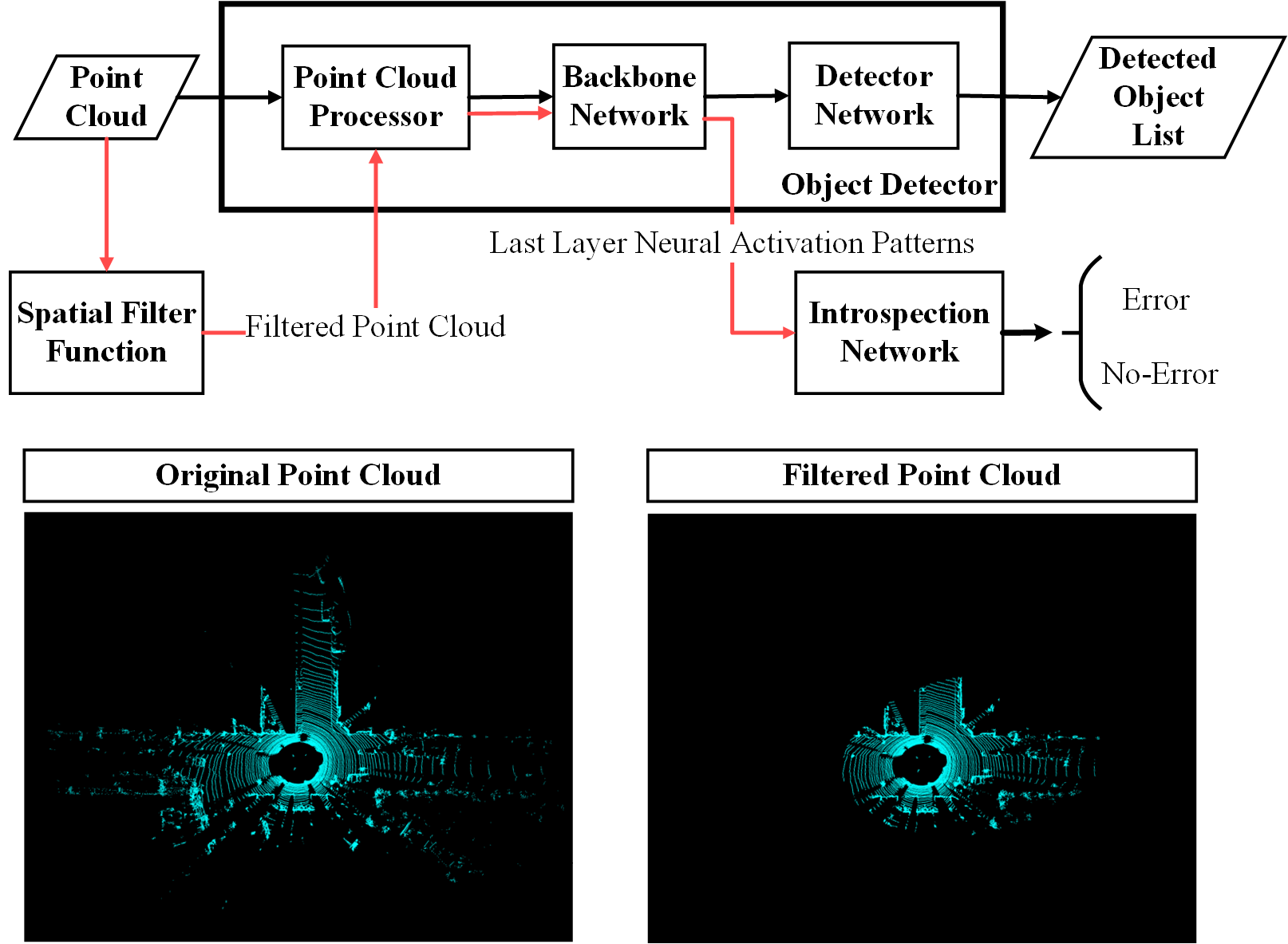}
    \caption{\small{LiDAR-based object detection pipeline depicted at the top starting with a processor network that extracts features from the point cloud. The extracted features are then processed by a backbone network to extract features and fed to the detector network. The introspection framework consists of the spatial filtering function, a copy of the point cloud processor, a copy of the backbone network, and the introspection network. The point cloud is first filtered and fed to the copy of the processor and backbone to extract neural activation patterns. The generated patterns are then fed into the Introspection Network, which classifies the collected point cloud as `Error' or `No-Error'. An example illustration of the original and the filtered point cloud is depicted at the bottom.}}
    \label{fig:overview}
\end{figure}

State-of-the-art (SOTA) object detection mechanisms utilise deep neural networks (DNNs) that have recently shown remarkable performance on various benchmarks~\cite{detectionsurvey,detectionsurvey2}. Despite these advancements,  object detectors in ADS are not error-free for a number of reasons. To begin with, the deployment conditions are so versatile that it is impossible to encompass all of them during training. In addition, the training process is data-dependent and stochastic, e.g., the neural network weights may be randomly initialised. In other words, some parts of the DNNs are non-deterministic by design. Furthermore, noise and input degradation can affect how well the DNN-based models work in practice~\cite{Eduardo2020, Buchholz}. Hence, detecting errors in DNN-based systems fundamentally differs from traditional automation systems (e.g., anti-lock braking system) and is hereafter defined as \textit{``introspection"}.

Recently, various methods for introspection in DNN-based systems have been explored, leveraging distinct input representations and techniques for identifying errors \cite{hakansurvey}. One such method is based on confidence or uncertainty, utilising the output's certainty level for error detection as discussed in \cite{dropoutod,feng2018towards,fengprobsurvey}. Another strategy involves identifying discrepancies (inconsistencies) by different systems running in parallel, exemplified by the comparison of the outputs provided by object detection and tracking in~\cite{failingtolearn}. For ADS in controlled settings, using a history-based introspection approach shows some promise. This method relies on the storage and retrieval of past experiences for error identification, as mentioned in \cite{fitforpurpose}. Additionally, predicting when system performance, indicated by metrics like mean average precision (mAP), falls below a specific standard in runtime helps in pinpointing errors, as detailed in \cite{rahmanper,hakaniccv,hakantiv}. %Additionally, predicting performance metric, like mean average precision (mAP), drops, where the system performance falls below a specific standard,  in runtime offers a way to pinpoint errors, as detailed in \cite{rahmanper,hakaniccv,hakantiv}.

\begin{figure}[t]
    \centering
    \includegraphics[width=\linewidth]{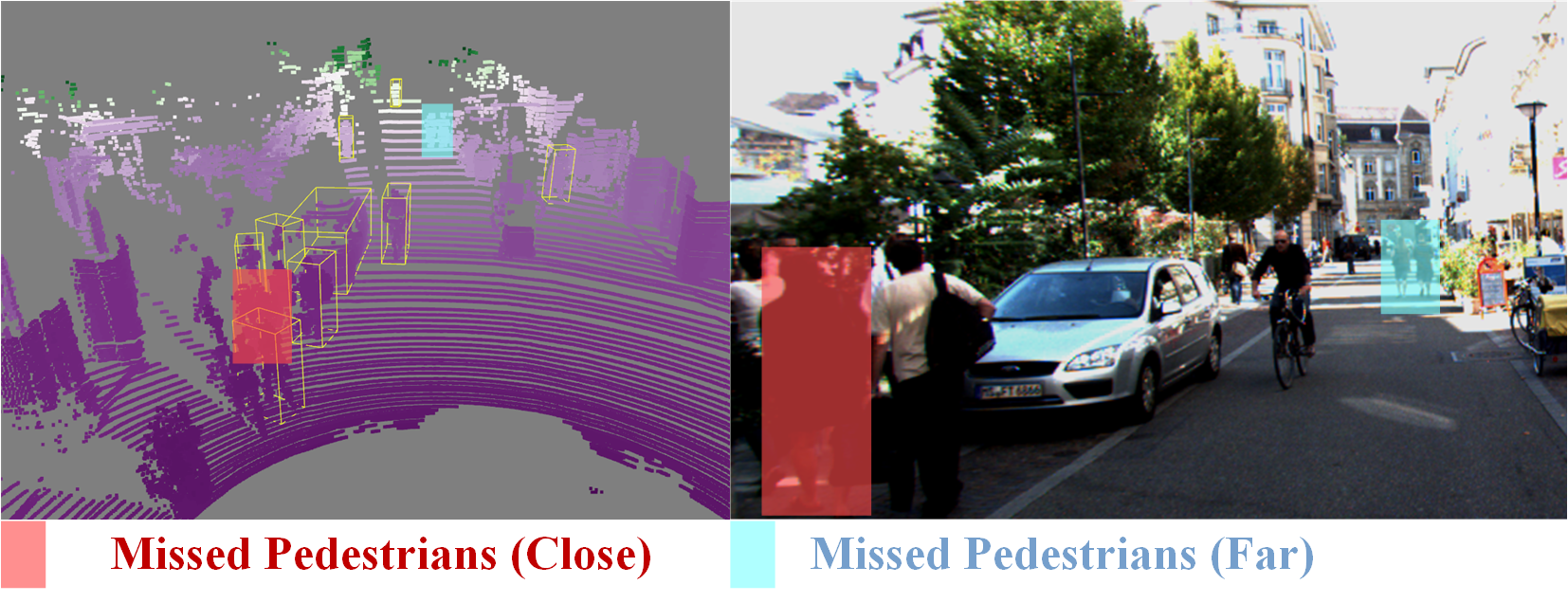}
    \caption{Point Cloud and Image from the KITTI Dataset \cite{kitti}, illustrating two pedestrians with different degree of criticality that were not detected by the 3D object detector. Coloured boxes are used to emphasise the correspondence of these objects between the point cloud and the image.}
    \label{fig:methodm}
\end{figure}

Despite gaining significant attention in the past decade, introspection of DNN-based mechanisms in ADS has been centred around 2D object detection and classification functions \cite{hakansurvey}. However, the three-dimensional (3D) nature of the world requires a comprehensive understanding of the environment in 3D to ensure the resilience and robustness of ADS applications. 
Additionally, in recent years, investigating neural activation patterns for either uncertainty estimation or for identifying performance drops has gained popularity due to its flexibility and ease of integration into other systems \cite{rahmanper,hakaniccv}. Despite that, only limited attempts for the monitoring of commonly utilised LIDAR-based 3D object detectors have been made so far, which are mainly clustered around uncertainty estimation~\cite{feng2018towards,feng2019can}. Furthermore, current research on introspecting the object detection process primarily focuses on frame-level analysis. % or generates outcomes for each detected object.
For example, in~\cite{hakancvpr}, the introspection system raises an alert if it is predicted that at least an object in the 3D point cloud hasn't been detected. Yet, from a safety point of view, not all objects in a scene hold equal significance (see \Cref{fig:methodm}). Clearly, the oversight of distant objects is less likely to lead to safety-critical situations, suggesting a need for a prioritisation mechanism in error detection.

To that end, our research introduces a distance-based spatial filtering mechanism for LIDAR-based point cloud data and leverages activation maps for introspection that are centred on the ego vehicle's immediate danger zone. 
%We apply filters to the LIDAR-based point cloud data and ground truth labels, directing the introspection network to focus on detecting missing objects in an area of interest around the ego vehicle. 
Filtering of the point cloud data may result in losing crucial scene information if some objects intersect the border of the spatial filter. To determine if vital information has been erased through this process, we also employ an alternative  introspection mechanism. That mechanism maintains the complete point cloud data while only implementing filtering of the ground truth labels, still allowing the introspection network to learn to pinpoint errors in the area of interest. 

For evaluating the performance of the developed introspection mechanism, we utilise well-known baseline object detection models in ADS, i.e., PointPillars\cite{pointpillars} and CenterPoint\cite{centerpoint}, and extract neural network activation patterns from the last layer of their backbone model, i.e., SECOND~\cite{SECOND}. An error is declared if the object detector misses an object in the area of interest, and the (binary) labels are paired with the raw neural activation patterns of the filtered point cloud. Subsequently, an introspector convolutional neural network (CNN) is trained on the generated pairs, hereafter referred to as the error dataset. A high-level graphical summary of our mechanism is presented in Fig.~\ref{fig:overview}. In summary, the contributions of this paper are:
\begin{itemize}
    \item A novel introspection mechanism for LIDAR-based point clouds is designed. It is based on the filtering of neural activation maps so that they focus only on the vicinity of the ego vehicle. % that generates  and learns from focused neural activation maps.
    \item We investigate the effect of information loss due to spatial filtering by comparing its performance with that of a label-only filtering mechanism.
    \item Extensive evaluations of the proposed mechanism in terms of error detection capability on two well-known public driving datasets and 3D object detectors.

\end{itemize}
%  In Section~\ref{sec:litr}, we review the current literature on introspection of object detection.

The structure for the rest of this paper is as follows: Section~\ref{sec:meth} outlines the proposed filtering mechanism and introspection method. Experimental settings and performance evaluation are presented in Section~\ref{sec:perfeval}. Finally, key takeaways of the paper are given in Section~\ref{sec:conc}.

\section{Proposed Method}\label{sec:meth}
\begin{figure*}[t]
    \centering
    \includegraphics[width=\linewidth]{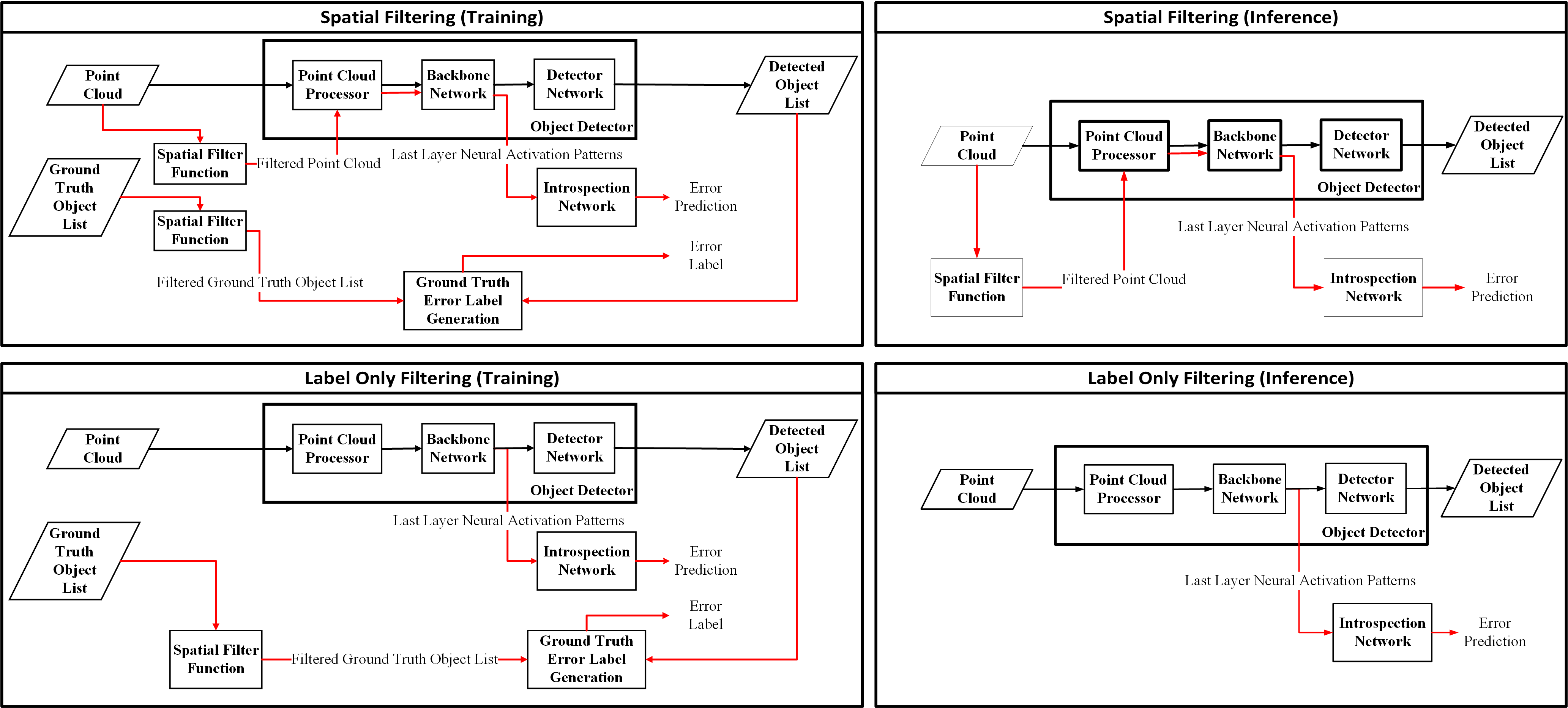}
    \caption{\small{Block diagrams for the introspection mechanism using spatial filtering (top) and label-only filtering (bottom) during training and inference. In spatial filtering, both the ground truth object list and the point cloud data are filtered during training. To process the filtered point cloud to extract activation maps, a copy of the network until the detector network is utilised within the proposed introspection mechanism. In label-only filtering, this copied network is not required, and only the ground truth labels are filtered for generating the error labels during training.}}
    \label{fig:methodc}
\end{figure*}
This section introduces a novel introspection mechanism with spatial filtering for 3D object detection, leveraging extracted activation maps from the object detector's backbone network. Unlike SOTA introspection studies, our approach addresses the problem of false negatives in the close vicinity of the ego vehicle to declare an error, as opposed to considering all objects in the input point cloud. Since distant objects usually don't pose an immediate threat to the ego vehicle and are often overlooked due to being hidden or because the point cloud data is not dense enough, we focus on spatially filtered point cloud data. This helps us identify features that are crucial for identifying missing objects near the ego vehicle.

One may notice that directly filtering the input point cloud may result in information loss. For example, if an object lies partially within the area of interest, spatial filtering would eliminate important information from the point cloud that can be useful to extract and pass onto the introspection network. To investigate the effect of this drawback on the introspection mechanism, we also adapt our mechanism to operate with the whole point cloud but only focus on the objects being (partially or fully) within the area of interest. 

Furthermore, SOTA introspection models for 2D object detection, such as those in~\cite{hakaniccv, hakantiv}, learn the relationship between activation patterns and mAP. However, mAP can be misleading for introspection when there are different classes of objects present in the frame~\cite{hakantiv}. For example, if the input frame contains multiple vehicles and a single pedestrian, where the pedestrian and majority of the vehicles are detected, the frame can still be labelled as no-error regardless of the missed vehicles' location. 
%This issue becomes critical in scenarios, where the area of interest is determined through spatial filtering. 
False negatives in such cases could lead to disastrous consequences when the missing object is located near the ego vehicle. Hence, in this paper, we opt to learn the relationship between activation patterns and missed objects (false negatives) within a part of the 3D point cloud. If at least one object within the area of interest is not detected, the frame is classified as `Error'.

In order to obtain some preliminary results on the effect of spatial filtering on the introspection model, the area of interest in this paper is assumed to be two-dimensional (2D) and  onto the horizontal plane. Several shapes, such as rectangles, triangles or ellipses, can be supported. To begin with, we choose an ellipse for the area of interest.
%To extract the area of interest, we introduce a 2D shape-based mechanism centred around the ego vehicle centre, or sensor position. 
Our mechanism does not consider the z-coordinate (height), and accepts any point of the LIDAR-based point cloud as long as their projection n the horizontal plane fall within the ellipse, see \Cref{fig:overview} for an example illustration. % due to its superior alignment with the geometrical characteristics of road objects, its efficient coverage capability, and its flexibility in adapting to various traffic scenarios.

Although the original ellipse is centred around the ego vehicle, we can introduce offsets in both longitudinal and lateral coordinates to allow shifting of the ellipse. In this study, we applied an offset of 5-10m in the longitudinal direction towards the front of the ego vehicle, as it is more significant for safety. Our empirical analysis showed that a filter with 50m coverage ([-20m,30m]) in the longitudinal axis and 20m ([-10m,10m]) in the lateral axis provides the best introspection results, where almost every frame has one or two objects in the selected area in both datasets. The selected values for the size of the ellipse are in accordance with the size of the coverage area of modern LIDAR sensors \cite{lidarrange}.

\subsection{Spatial Filtering}
Once the parameters of the filter and the shape are set, the introspection mechanism is ready for training. As shown in \Cref{fig:methodc}, in both training and inference, spatial filtering removes the part of the point cloud that lies outside the area of interest. The filtered point cloud is then fed to the copy of the object detector's point cloud processor and backbone, where the introspection network uses the output at the last layer of the backbone to identify detection errors. 

To train, we use the bounding boxes from the ground truth object list and discard those boxes which fall entirely outside the area of interest. To do that, we determine the corners of each box, verifying if any point of it lies within the area of interest. As a result, the filtered ground truth list contains boxes that may fall partially or entirely within the area of interest. We then compare  the filtered ground truth with the detected object list to identify possibly undetected objects. An object in the filtered ground truth list is considered to be detected if its intersection over union (IoU)  with any object from the detected object list is higher than 0.7. Hence, if an object in the filtered ground truth list does not match any of the detected objects, a detection error is declared. During inference, as no ground truth information is available, we remove obtaining and filtering the ground truth object information from the mechanism and apply only filtering on the point cloud data.

\subsection{Label-only Filtering}
\label{sec:Label-only}
Label-only filtering is employed to assess whether cropping the point cloud during spatial filtering is detrimental to introspection. This mechanism uses exactly the same steps with spatial filtering to create the 'error' and 'no error' labels during training, but it does not filter the input point cloud. The introspection model is expected to concentrate on the area of interest, guided by the error labels, not by a filtered point cloud.
%the ground truth bounding boxes, discarding those that lie entirely outside the area of interest. One may see in \Cref{fig:methodc} that the way is exactly the . As shown , unlike spatial filtering, . 
During inference, this mechanism, similar to spatial filtering, lacks ground truth object information and, therefore, does not employ filtering. It operates as if no filtering is  applied, similar to the method in~\cite{hakancvpr}. In some sense, this mechanism is lighter in terms of computation and easier to integrate into the perception system because it doesn't require a copy of the point cloud processor and backbone network.

\section{Performance Evaluations}\label{sec:perfeval}
This section presents the experimental setup and performance evaluation on comparing introspection mechanisms utilising neural activation patterns with spatial area of interest filtering. Before that, we justify the selection of object detectors, driving datasets, and the adaptation of the SOTA introspection mechanism and key performance indicators used in our analysis.

\subsection{Object Detectors}
We investigate the behaviour of introspection systems on 3D object detection using two popular models. First, we utilise the PointpPillars model~\cite{pointpillars}, a widely used baseline model in 3D object detection in ADS \cite{autowareauto}. PointPillars is a common baseline model for 3D object detection, particularly tailored for LIDAR data, which proposes a novel encoder architecture  transforming the irregular and sparse 3D point clouds generated by LIDAR sensors into a structured format called ``pillars". These pillars are essentially vertical columns that capture the points in a defined columnar space, simplifying the complexity of 3D data processing. Once the data is organised into pillars, PointPillars employs a neural network to learn distinctive features from each pillar. This model projects the learned features onto a pseudo-image, enabling the use of a 2D CNN for further processing. Despite the recent advancements in the domain, PointPillar model still used in earlier software stack for automated driving \cite{autowareuniverse} and remains a baseline for 3D object detection in ADS.
    
Additionally, for a comprehensive evaluation of introspection in recent detection mechanisms, we employ the CenterPoint model. This model focuses on identifying the centre of objects first and then regresses to define the bounding box. This method contrasts with other detectors that directly regress the corners of the bounding box. The fundamental motivation behind CenterPoint is the observation that the centres of objects remain invariant to rotation. This characteristic ensures reliable detection even when vehicles assume different orientations due to varying road conditions. In addition, this model is utilised in Autoware Foundation's updated software stack ``Autoware.Universe" \cite{centerpoint}. 

In terms of implementation and training of these models, we have employed OpenMMDet3D framework \cite{mmdet3d} and utilised the pre-trained models on KITTI \cite{kitti} and NuScenes\cite{nuscenes} datasets. In this framework, both models use a network called SECOND \cite{SECOND}, which utilises sparse convolution operations to provide faster operation with the sparsity of the LIDAR's point cloud data.

\subsection{Driving Datasets}
To extract and generate the error dataset, we have utilised two widely-used datasets based on their use in both introspection and ADS domains: KITTI~\cite{kitti} and NuScenes~\cite{nuscenes}. 

The KITTI dataset consists of over 14,000 annotated images captured by a camera and a Velodyne LiDAR mounted on a car driving through urban environments in Karlsruhe, Germany. The training set contains 7,481 samples, including annotated point clouds, and the test set includes 7,518 samples but no annotations. The benchmarking is typically done using only three classes: car, pedestrian, and cyclist. Additionally, It is important to note that the labelling on the data only covers objects in front of the vehicle.%, which is also reflected in the training of PointPillar's model, which is depicted in Figure~\ref{fig:activationkittinput}, in Section~\ref{sec:qualitative}

The NuScenes dataset is a comprehensive collection comprising over 1,000 diverse driving scenes captured across various urban locations in Boston, USA, and Singapore. This dataset is significantly richer in annotations than many of its counterparts, featuring six camera feeds, five RADAR, and a full 360-degree LIDAR sweep. In total, the NuScenes dataset includes 1.4 million images, 390k LIDAR sweeps, and 1.4 million 3D bounding box annotations across 23 object classes. The dataset is split into a training set, a validation set, and a test set. Additionally, the dataset provides detailed annotations not just for objects in front of the vehicle but in its entire surroundings, offering a 360-degree perspective crucial for training and evaluating advanced object detection models like CenterPoint. 

\subsection{Introspection Training and Implementation Details}
To train the introspection network, we have utilised stochastic gradient descent (SGD) optimiser, with focal loss function \cite{focal}. Additionally, due to the imbalance in the error datasets, we have calculated class weights \cite{classweight} using training data and fed to the loss calculation along with the gamma ($\gamma$) value of five to mitigate the issue. We also implemented an early stop mechanism with a patience setting of 15 epochs, coupled with a learning rate scheduler that scales down the learning rate by a factor of 0.7 after a patience period of 10 epochs. All networks were trained for a total of 200 epochs using this approach. We experimented with learning rates of 0.01, 0.001, and 0.0005, among which 0.01 yielded the best performance. The batch size for this training was set at 64. Additionally, for neural network development and training PyTorch and Torchvision are utilised. Detection evaluation and metric calculation are done with Torchmetrics. Complexity calculation and inference time calculation are done with the time library in Python. Finally, all experiments are done on a machine equipped with an Intel(R) Core(TM) i9-10980XE CPU and NVIDIA RTX 3090 GPU.

\subsection{Performance Metrics}
Since the decision to use the introspection method is based on binary classification for all models considered in this paper, the following metrics are selected to evaluate the performance. % of the selected baselines.
\begin{itemize}
\item \textbf{Area Under Receiver Operating Characteristic Curve (AUROC):} Provides an indicator of how well a classifier distinguishes between the positive ('error') and negative ('no-error') classes. It measures the model's ability to avoid false classifications, with a higher AUROC indicating better performance.

\item \textbf{Recall (Positive and Negative):} Measures the classifier's ability to correctly identify true positives and true negatives. Positive Recall (also known as \textit{Sensitivity} or True Positive Rate) quantifies the proportion of actual positives correctly identified by the model. Negative Recall (also known as \textit{Specificity} or True Negative Rate) quantifies the proportion of actual negatives that are correctly identified. High recall values for both positive and negative classes indicate a model's effectiveness in correctly classifying both error and no-error instances.
\end{itemize}

\subsection{Performance Comparison}\label{sec:perfcomp}
\begin{table}[t]
    \centering
    \begin{tabular}{ccccc}
    \toprule
         \makecell{\textbf{Dataset} /\\\textbf{Method}}&\textbf{ Input}&\textbf{Rec.$_{(-)}$}&\textbf{Rec.$_{(+)}$}&\textbf{AUROC} \\\midrule
         % \multirow{5}{*}{\makecell{Kitti / \\ PointPillars} }&\makecell{SF}&0.1479&0.9408&0.6000\\
         % &\makecell{RA}&0.6268&0.8105&0.8036\\
         \multirow{3}{*}{\makecell{KITTI / \\ PointPillars} }&\makecell{Label-only Filtering}&\textbf{0.8077}&\textbf{0.7074}&{\textbf{0.8347}}\\
         &\makecell{Proposed with SF}&0.9649&0.0625&0.6193\\
         &\makecell{Proposed with RA}&\underline{0.6368}&\underline{0.7091}&\underline{0.7384}
\\\midrule

         % \multirow{5}{*}{\makecell{NuScenes / \\ CenterPoint}} &\makecell{SF}&0.2607&0.9217&0.7322\\
         % &\makecell{RA}&0.7123&0.8581&0.8919\\
         \multirow{3}{*}{\makecell{NuScenes / \\ CenterPoint}} &\makecell{Label-only Filtering}&\underline{0.7883}&\underline{0.6207}&\underline{0.7701} \\
         &\makecell{Proposed with SF}&0.4819&0.8067&0.7317\\
         &\makecell{Proposed with RA}&\textbf{0.5330}&\textbf{0.8646} &\textbf{0.8092} \\\bottomrule
    \end{tabular}
    \caption{\small{The performance of the proposed introspection model using raw activations (RA) with spatial and label-only filtering is evaluated. The performance using statistical features (SF) combined with spatial filtering is also presented. SF results using label-only filtering are not reported due to poor performance. Metrics: Recall for negative (Rec.${(-)}$) and positive classes (Rec.${(+)}$), and overall AUROC. Top AUROC model is in bold, second-best underlined.}}
    %Error detection performance of introspection models with using raw activations (RA) and statistical features (SF) \cite{rahmanper} with proposed spatial filtering, on KITTI (using PointPillars detector) and NuScenes (using CenterPoint detector). Metrics include Recall for negative (Rec.${(-)}$) and positive classes (Rec.${(+)}$), and AUROC for overall classification capability. The best-performing model based on AUROC is highlighted in bold, and the second-best is underlined. Label-only filtering is not applied with SF mechanism due to its poor performance on other settings.}
    \label{tab:comparison}
\end{table}
In this section, we present the evaluation of the introspection mechanism with spatial filtering of the neural activation maps for 3D object detection in ADS. We also investigate whether such filtering results in substantial information loss for the purpose of introspection. To do so, we utilise activation maps generated from non-filtered point cloud data but still focus solely on objects within the designated area of interest to determine if the input frame is an error or not. This has been referred to as `Label-only Filtering' in Section~\ref{sec:Label-only}. This approach enables an assessment of whether access to full scene information brings any benefits.  A high-level summary of how both mechanisms operate in the training process is presented in \Cref{fig:methodc}. Finally, we employ a statistical feature (SF) extraction mechanism, adapted from 2D object detection introspection \cite{rahmanper}, as a baseline. This method employs statistical functions—mean, max, standard deviation—combined with global pooling on 3D activation maps to generate 1D vectors, which are subsequently concatenated and given to the introspection network. We adapt this mechanism to use spatial filtering to extract statistical features from filtered activations. 

It is also essential to mention that the proposed mechanism is not compared with the no-filtering-based introspection model~\cite{hakancvpr}, as the main challenges tackled by  the two mechanisms are significantly different. To clarify, in no-filtering-based introspection, the aim is to classify the whole scene with a missed object or a set of missed objects depending on the tolerance ratio, while in this study, we aim to identify the missed objects in the close vicinity of the ego vehicle determined by spatial filtering and tolerate other missed objects. Next, we discuss how the distribution of `error' and `no error' samples changes when filtering is applied to highlight the importance of spatial filtering. To clarify, in the KITTI dataset, 63\% of the samples are labelled as errors when no filtering is applied, while the filtering mechanism shows that only 33\% of the samples have errors in the vicinity of the ego vehicle. Similarly, a significant change is observed in the more diverse and complex NuScenes dataset. No filtering method labelled 98\% of the samples as errors highlighting that in almost all samples, the detector missed at least one object. However, the ratio of error samples has reduced to 75\% with spatial filtering. Given that detection errors outside the area of interest are likely insignificant from a safety point of view, spatial filtering will reduce the number of unnecessary perception system alerts improving the driver's comfort in Level 3 autonomy.

\subsubsection{Error Detection Performance}
As presented on \Cref{tab:comparison}, the proposed method filtering the raw activation patterns provides the best result in terms of AUROC and error detection recall in the NuScenes dataset, and provides competitive results against label-only filtering in KITTI dataset. On the contrary, the approach using SF after filtering is not sufficient for error detection. This may be due to the channel-wise over-simplification of 3D activation maps, which may work well in densely activated neural patterns of image-based 2D object detection, but fails to adapt in 3D point clouds. \Cref{tab:comparison} also shows that utilising activation maps from the full scene is not better than the proposed filtering in more diverse and complex datasets, NuScenes. % as our proposed mechanism precedes the use of full point cloud in terms of AUROC. 
This may be due to the sparsity of the point cloud and lack of information in faraway areas, which, in the end, also confuse the introspection model. In the KITTI dataset, label-only filtering demonstrates superior AUROC performance despite exhibiting similar recall rates in identifying the `error' class as compared to the spatial filtering mechanism. This enhanced performance of the label-only mechanism may be attributed to the fact that the KITTI dataset utilises only half of the scene, thereby having a more limited focus area. Also, we see that when full point cloud information is used, the model tends to learn better features about the `no-error' class, resulting in reduced recall in the positive class and increased recall in the negative class.

\subsubsection{Activation Visualisation}
\begin{figure}
    \centering
    \includegraphics[width=\linewidth]{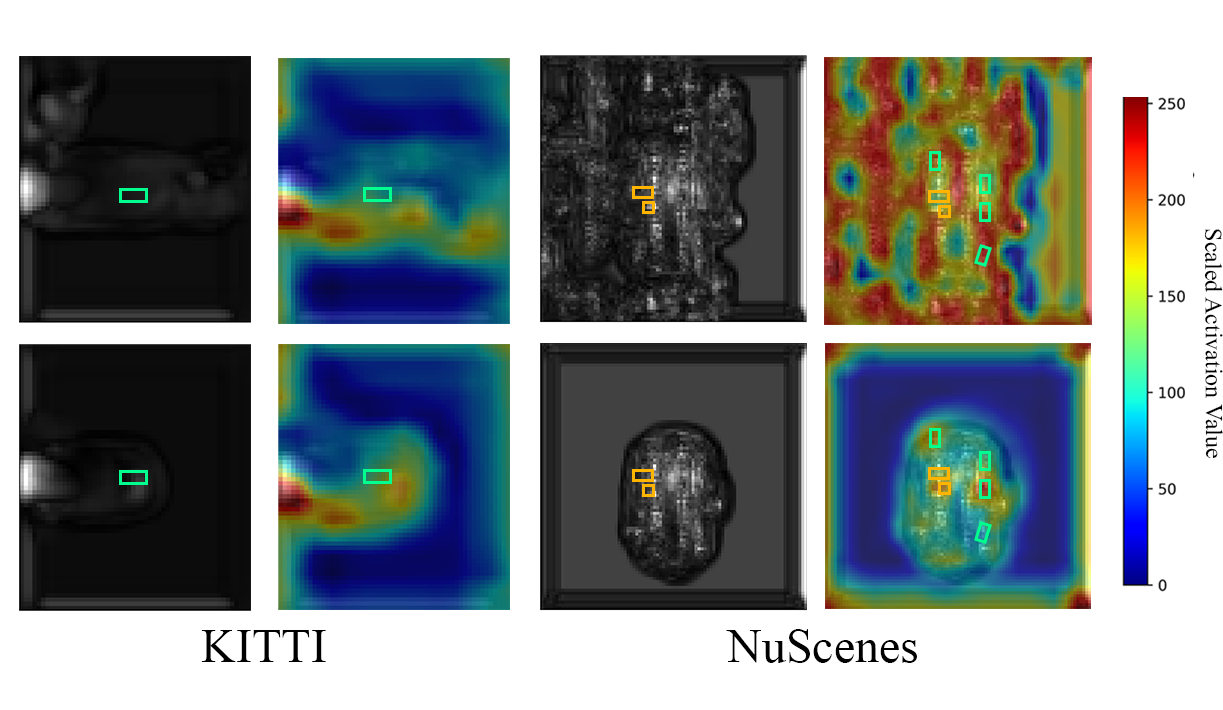}
    \caption{\small{Max activation maps and Eigen-CAM visualisations, for example, frames on the KITTI and NuScenes datasets. The top row indicates the results of label-only filtering, while the bottom shows our proposed spatial filtering mechanism. The first and third columns display the channel-wise max activations for the KITTI and NuScenes datasets, while the second and fourth columns exhibit the respective Eigen-CAM heatmaps that highlight areas critical to the classification. For clarity, objects correctly detected are marked with green boxes, while missed ones are highlighted with orange boxes.}}%Activation maps and Eigen-CAM visualisations for KITTI and NuScenes datasets: top row shows label-only filtering, bottom row spatial filtering. Columns one and three present max activations, two and four show Eigen-CAM heatmaps. Correct detections are marked green, missed ones orange.}
    \label{fig:activis}
\end{figure}
We present the visualisation of early layer activations of the proposed introspection model to provide a deeper understanding of the decision-making of the models considered in this paper. % of different layers extracted from object detector's backbone, and our proposed mechanism. 
For this purpose, we have extracted the first layer activation maps of the introspector's CNN, i.e. ResNet 18, and applied a well-known class-activation map generation method, Eigen-CAM \cite{eigencam}, which simply calculates the principal components of the activation maps and generates visualisations highlighting the activated regions. We opt to utilise the first layer as the later layers have low resolution, which may result in oversimplified class activations. The resulting visualisations are presented for each mechanism and for both datasets in \Cref{fig:activis}. The driving direction is from left to right in KITTI and from bottom to top in NuScenes. It is also important to recall the distinction between the two  datasets: While KITTI focuses on objects in front of the vehicle, NuScenes provides a comprehensive 360-degree view. % in its processed point cloud.

In the KITTI dataset (first and second columns), a no-error instance is showcased, exemplifying the precise detection of a vehicle situated ahead of the ego vehicle. That is also the only object in the scene belonging to the pedestrian or vehicle classes. As presented in the \Cref{fig:activis}, the activations are more focused in front of the ego vehicle as opposed to the label-only filtering mechanism where activations are extracted from the full point cloud. This difference may support that the introspection model can be confused by other objects outside the area of interest, as features are also extracted and seen by the introspector for the full frame in the label-only mechanism. 

In the NuScenes dataset (third and fourth columns), an error case is given with missed detections of a pedestrian and a vehicle behind the ego vehicle, which all introspection models correctly identify. Even though the scene contains around 40 objects belonging to the vehicle or pedestrian classes, the introspection mechanisms are capable of predicting undetected objects. This indicates the effectiveness of using activation maps for identifying errors in LIDAR-based 3D object detection. Compared to KITTI, activations are confined to the area of interest for the spatial-filtering mechanism, as shown in the extracted activation maps. Conversely, in the label-only mechanism, the network attends to multiple scene parts, including regions beyond the area of interest, potentially leading to confusion in determining scene errors.

\section{Summary \& Conclusion}\label{sec:conc}
In this research, we investigated the impact of a focused activation map generated with spatial filtering on the error detection performance of 3D object detection in automated driving systems (ADS). 
%Prior studies have typically focused on the final layer neural activations in 2D object detection, often finding earlier layer activations less effective. However, 
We hypothesised that in the context of ADS, not all objects are equal regarding safety criticality. Hence, the error detection mechanism should focus on the close vicinity of the ego vehicle. Consequently, an error detection mechanism prioritising the immediate vicinity of the ego vehicle using spatial filtering is proposed. We utilise the KITTI and NuScenes datasets, applying the proposed filtering to both labels and point cloud data. We then test the effect of the filtering  using PointPillars and CenterPoint models with the extracted activation patterns to identify scenes with false negatives.

Our findings reveal that the proposed spatial filtering mechanism provides good overall results as well as error detection performance. We also show that, although information loss occurs in our proposed mechanism, the introspection is still able to perform well compared to the case where no point-cloud filtering is applied. %Lastly, we show that the overhead added with our mechanism is still within the real-time working range for ADS applications.
% Further analysis showed that while earlier layer activations increased the model's confidence in correctly classified samples in large and diverse datasets, they decreased confidence in smaller datasets. 
%We also observed that although all methods are viable for real-time inference, relying on earlier layer activations demands  
Since introspection in 3D object detection, particularly in ADS, is a relatively unexplored subject, future studies are needed on both introspection and spatial filtering. They should focus on developing new mechanisms utilising additional information available in ADS for both introspection and filtering. Additionally, it is essential to investigate the tolerance level for a missed object with spatial filtering. Similarly, although this study mentions empirical testing on the best range for filtering, different filtering shapes and ranges should be investigated. Lastly, it is essential to develop metrics that consider the objects' distance, heading, and speed to the error metric for more realistic and robust error detection.

\bibliographystyle{IEEEtran}
\bibliography{refs}

\addtolength{\textheight}{-12cm}   % This command serves to balance the column lengths
                                  % on the last page of the document manually. It shortens
                                  % the textheight of the last page by a suitable amount.
                                  % This command does not take effect until the next page
                                  % so it should come on the page before the last. Make
                                  % sure that you do not shorten the textheight too much.
\end{document}